\pdfoutput=1

\documentclass[11pt]{article}

\usepackage[final]{acl}

\usepackage{times}
\usepackage{latexsym}

\usepackage[T1]{fontenc}

\usepackage[utf8]{inputenc}

\usepackage{microtype}

\usepackage{graphicx}

\usepackage[utf8]{inputenc}     
\usepackage[T1]{fontenc}        
\usepackage{hyperref}           
\usepackage{url}                
\usepackage{booktabs}           
\usepackage{amsfonts}           
\usepackage{nicefrac}           
\usepackage{microtype}          

\usepackage{colortbl}
\usepackage{booktabs}
\usepackage{multirow}
\usepackage{graphicx}
\usepackage{CJKutf8}
\usepackage{tgcursor}
\usepackage{bm}
\usepackage{arydshln}
\usepackage{amsfonts}
\usepackage{comment}
\usepackage{subcaption}
\usepackage[font=small,labelfont=bf]{caption}
\usepackage{float}
\usepackage{babel}
\usepackage{wrapfig}
\usepackage{tikz}
\usepackage[xcolor,framemethod=tikz]{mdframed}
\usepackage{markdown}

\usepackage{zi4}

\definecolor{backgray}{rgb}{0.92,0.92,0.92}
\definecolor{textgray}{rgb}{0.6,0.6,0.6}
\definecolor{myblue}{HTML}{d4e1f5}
\definecolor{mygreen}{HTML}{CAE5CD}
\definecolor{myred}{HTML}{ffcccc}
\definecolor{eqblue}{HTML}{48bbdb}
\definecolor{eqgreen}{HTML}{3C8031}
\definecolor{eqred}{HTML}{EE2967}
\definecolor{eqyellow}{HTML}{FFDF42}
\definecolor{eqorange}{HTML}{F26035}
\definecolor{aaagreen}{HTML}{C3FFC5}
\definecolor{aaared}{HTML}{FFC3D1}
\definecolor{aaablue}{HTML}{C3D0FF}

\DeclareRobustCommand{\aaagreen}[1]{\tikz[baseline=(X.base)]{\node(X)[rectangle, fill=aaagreen, rounded corners, text height=.8ex,text depth=-0.5ex]{\textit{#1}};}}
\DeclareRobustCommand{\aaared}[1]{\tikz[baseline=(X.base)]{\node(X)[rectangle, fill=aaared, rounded corners, text height=.8ex,text depth=-0.5ex]{\textit{#1}};}}
\DeclareRobustCommand{\aaablue}[1]{\tikz[baseline=(X.base)]{\node(X)[rectangle, fill=aaablue, rounded corners, text height=.8ex,text depth=-0.5ex]{\textit{#1}};}}

\newcommand{\ja}[1]{
    \begin{CJK}{UTF8}{ipxm}
    #1
    \end{CJK}
}

\newcommand{\mybenchmark}{\textsc{AdTEC}}
\newcommand{\aaa}{A$^3$}
\newcommand{\taskaccept}{\textsc{Ad Acceptability}}
\newcommand{\taskconsist}{\textsc{Ad Consistency}}
\newcommand{\taskaaa}{\textsc{\aaa~Recognition}}
\newcommand{\tasksim}{\textsc{Ad Similarity}}
\newcommand{\taskpe}{\textsc{Ad Performance Estimation}}
\newcommand{\calm}{CALM2$_\mathrm{7b}$}
\newcommand{\elyza}{ELYZA$_\mathrm{7b}$}

\mathchardef\mhyphen="2D

\title{\mybenchmark: A Unified Benchmark for Evaluating Text Quality\\in Search Engine Advertising}

\author{
 \textbf{Peinan Zhang\textsuperscript{\scriptsize $\spadesuit$}},
 \textbf{Yusuke Sakai\textsuperscript{\scriptsize $\diamondsuit$}},
 \textbf{Masato Mita\textsuperscript{\scriptsize $\spadesuit$}},
 \textbf{Hiroki Ouchi\textsuperscript{\scriptsize $\spadesuit$$\diamondsuit$}},
 \textbf{Taro Watanabe\textsuperscript{\scriptsize $\diamondsuit$}}
\\[0.1cm]
 \textsuperscript{\scriptsize $\spadesuit$}CyberAgent,
 \textsuperscript{\scriptsize $\diamondsuit$}Nara Institute of Science and Technology
\\
 \texttt{\{zhang\_peinan,mita\_masato\}@cyberagent.co.jp}\\
 \texttt{\{sakai.yusuke.sr9,hiroki.ouchi,taro\}@is.naist.jp}
}

\begin{document}
\maketitle
\begin{abstract}
As the fluency of ad texts automatically generated by natural language generation technologies continues to improve, there is an increasing demand to assess the quality of these creatives in real-world setting.
We propose \mybenchmark, the first public benchmark to evaluate ad texts from multiple perspectives within practical advertising operations.
Our contributions are as follows: (i) Defining five tasks for evaluating the quality of ad texts, as well as constructing a Japanese dataset based on the practical operational experiences of advertising agencies, which are typically maintained in-house. (ii) Validating the per-formance of existing pre-trained language models (PLMs) and human evaluators on this dataset. (iii) Analyzing the characteristics and providing challenges of the benchmark.
Our results show that while PLMs have a practical level of performance in several tasks, humans continue to outperform them in certain domains, indicating that there remains significant potential for further improvement in this area. The dataset is publicly available at: \texttt{\url{https://cyberagentailab.github.io/AdTEC}}.
\end{abstract}

\section{Introduction}\label{sec:intro}

\begin{figure}[t]
    \centering
    \includegraphics[width=0.4\textwidth]{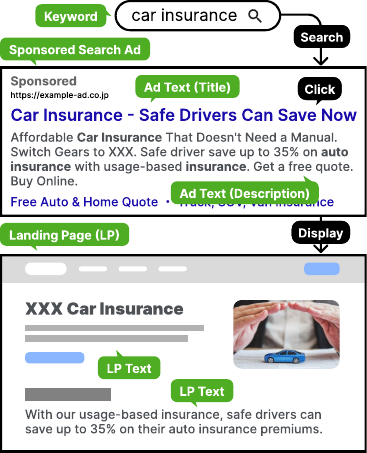}
    \caption{Overview of sponsored search ad and its key terms.}
    \label{fig:sponsored-search-ad}
\end{figure}

\begin{figure*}[t]
    \centering
    \includegraphics[width=0.92\textwidth]{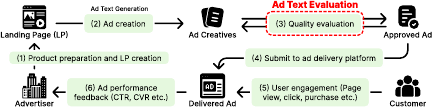}
    \caption{Generalized AdOps workflow described in \S\ref{sec:adops}: (1) The advertiser creates an LP to promote a product. (2) Based on the product information in the LP and target customers, text and graphics are designed by creators. (3) The creatives are evaluated based on fluency, attractiveness, regulations, legality, and other factors. (4) Once the creatives pass the quality evaluation, they are submitted to a delivery platform. (5) Customers respond to the displayed ads, such as page views, clicks, and purchases. (6) Based on the customer engagement, ad performance is reported back to the advertiser, and Steps 1-5 are repeated to improve the quality of the LP and ads.}
    \label{fig:adops}
\end{figure*}

Online advertising, especially sponsored search advertising (Figure \ref{fig:sponsored-search-ad}), is a dominant sector for vendors to promote their products, and the market size is estimated to grow by billions of dollars over the next few years~\cite{murakami2023survey}.
To meet the increasing demands of advertising operations (AdOps), such as creating ad texts from product information (Step 2 in Figure \ref{fig:adops}), the remarkable success of natural language generation (NLG) by pre-trained language models (PLMs)~\cite{Dong2021ASO,Vaswani2017AttentionIA,brown2020language,touvron2023llama} has given a boost to practical applications~\cite{hughes2019adnlg,kamigaito2021empirical,golobokov2022deepgen}, making advertising a huge industrial use case for natural language processing (NLP).


In this paper, we focus on evaluating the quality of ad texts generated by such models.
We refer to this process as \textit{ad text evaluation}, as depicted in Step 3 of Figure \ref{fig:adops}.
Ad text evaluation is crucial because low-quality ad texts , which may lack fluency, present inappropriate appeals, or convey misleading representations, can negatively impact advertisers.
Since it is costly and not scalable to have humans verify the quality of each text in high-volume domains, such as sponsored search advertising, there is a high demand for the development of automated quality estimators for ad texts.
The quality has multiple dimensions, such as appropriate wording, effective appeals, consistency between ad text and product information, and high predicted performance.
Although these dimensions should be included when evaluating automatic quality estimators, no benchmark currently exists to assess them.
Consequently, the bottleneck lies in verifying the quality of ad texts despite the ability to generate numerous creatives automatically, which hinders delivery volume scalability.
Thus, we aim to construct a benchmark to evaluate the quality of ad texts.

The primary challenge in constructing an ad text evaluation benchmark is the absence of a clear definition for tasks~\cite{murakami2022aspect,mita-etal-2024-striking}.
The lack of domain knowledge in AdOps complicates the understanding of high-quality ad text standards and the accurate definition of tasks.
However, the AdOps workflow is complex, relies on various platforms and formats, and encompasses multiple methodologies and metrics.
In addition, only a few companies possess the expertise to operate online advertising at scale.
Owing to legal and contractual obligations, advertising workflows and data are predominantly managed in-house, leading to a lack of publicly available datasets.
This scarcity makes it challenging to systematically reproduce and validate diverse methodologies in academia.
Consequently, research activity is limited in the advertising field, which leaves potential issues unaddressed and delays the application and development of cutting-edge technology.

To address these challenges, we propose \textbf{\mybenchmark}, the first public benchmark that defines and unifies tasks based on generalized AdOps workflows. Our major contributions are as follows:

\makeatletter
\renewenvironment{description}%
   {\list{}{\leftmargin=10pt 
            \labelwidth\z@ \itemindent-\leftmargin
            \let\makelabel\descriptionlabel}}%
   {\endlist}
\makeatother

\begin{description}
    \setlength{\parskip}{0pt} 
    \item[The First Public Dataset on Ad Text Evaluation.]
    We organized real-world advertising workflows and carefully designed five tasks to evaluate the quality of ad texts. This was based on the practical operational experiences of advertising agencies, using Japanese data as a case study.
    We are releasing this dataset\footnote{\texttt{\url{https://cyberagentailab.github.io/AdTEC}}}, marking it as the first publicly available dataset for ad text evaluation, as such data are typically maintained in-house and difficult to obtain.
    \item[Benchmark Experiments.] We validated the performance of existing PLMs, such as BERT, RoBERTa, and large language models (LLMs), as well as human evaluators on our proposed benchmark.
    \item[Dataset Analysis.] We analyzed the characteristics and identified potential issues of the dataset through experiments, demonstrating that our benchmark is challenging and highlighting potential areas for improvement and future research.
\end{description}

\begin{table*}[t]
\centering
\resizebox{\textwidth}{!}{%
    \begin{tabular}{@{}ccc@{}}
    \toprule
    \textbf{Ad text} & \textbf{LP text} &  \textbf{\textsc{Ad Accept. / Consist.}} \\
    \midrule
    Condominium Sales / Free Assessment Now & Let XX assess your single-family home! &  \texttt{acceptable} / \texttt{inconsistent} \\
    \cmidrule{1-3}
    Engineer's career / Engineer's Job & Find jobs at website XX! &  \texttt{unacceptable} / \texttt{consistent} \\
    \toprule
    \end{tabular}%
}
\caption{Examples of \taskaccept~and \taskconsist~tasks. Note that LP texts are only used in the \taskconsist~task.}
\label{tab:examples123}
\end{table*}

\section{Understanding Sponsored Search Advertising and its Workflow}\label{sec:adops}

In sponsored search advertising, titles and descriptions of ads that are relevant to keywords entered into search engines by customers appear as part of the search results.
As shown in Figure \ref{fig:sponsored-search-ad}, when a customer clicks an ad's URL, they are directed to a web page known as a landing page (LP).
The LP contains texts and images related to the advertiser's products, prompting the customer to view the product or make purchase decisions.

Operating such ads requires high expertise, owing to the variety of delivery platforms, properties, and formats.
To gain insights, we interviewed two types of experts familiar with AdOps: those overseeing AdOps departments at ad agencies and those directly involved in on-site operations.
Through these interviews, we explored the intricacies of AdOps, generalized the workflow, and delineated six steps, as illustrated in Figure~\ref{fig:adops}.
In this work, we focus mainly on Step 3 of quality evaluation.

\section{Constructing \mybenchmark}\label{sec:task-design}

Our goal is to develop a benchmark that captures multi-dimensional aspects of quality that are relevant to practical scenarios.
In addition, these aspects will be carefully curated to provide value for both real-world applications and various research purposes.

\subsection{Task Design}
We followed the workflow outlined in \S\ref{sec:adops} and asked the same experts from the interview to identify crucial parts within Step 3 in Figure \ref{fig:adops}.
Based on their insights, we generalized and defined tasks that adhere to the principle of evaluating ad texts either directly or indirectly.
\textit{Direct} evaluation tasks are used to test texts against strict criteria, such as a binary pass/fail outcome or numerical score to quantify the text quality.
These tasks serve as a checklist to ensure that minimum delivery standards are met.
\textit{Indirect} evaluation tasks are used to assist human evaluators in reviewing or refining texts or serve as a bridge to connect downstream tasks.
Based on the above principles, we designed five tasks: three direct evaluation tasks (\taskaccept, \taskconsist, and \taskpe) and two indirect evaluation tasks (\taskaaa~and \tasksim).\footnote{Note that all examples in this paper are translated from Japanese into English for presentation brevity.}

\begin{figure*}[t]
    \centering
    \begin{tabular}{cc}
        \begin{minipage}{0.45\textwidth}
            \centering
            \resizebox{\textwidth}{!}{%

                \begin{tabular}{@{}ll@{}}
                    \toprule
                    \textbf{Field}& \textbf{Example value}\\
                    \midrule
                    \multirow{1}{*}{\textbf{Title 1}} & {[}No.1{]} Card loan comparison site \\[0.1cm]
                    \multirow{1}{*}{\textbf{Title 2}} & A must-see for those in a hurry! \\[0.1cm]
                    \multirow{1}{*}{\textbf{Title 3}} & Instant Loan Secure Card Loan \\[0.2cm]
                    \multirow{3}{*}{\textbf{Desc. 1}} & The best place to get a card loan without \\
                                                      & telling anyone. You only need a driver's \\
                                                      & license to apply \\[0.2cm]
                    \multirow{3}{*}{\textbf{Desc. 2}} & Convenient to use ATMs at convenience \\
                                                      &  stores. Convenient and quick loans are  \\
                                                      & available if you apply before 10:00 p.m. \\[0.2cm]
                    \textbf{Keyword}                  & card loan \\[0.1cm]
                    \textbf{Industry}                 & finance \\[0.1cm]
                    \textbf{Score}                    & 82.3 \\
                    \bottomrule
                \end{tabular}%
            }
            \captionof{table}{Examples of \taskpe~task. Desc. represent Description. Score is the label, and others are inputs.
            }
            \label{tab:example-pe}
        \end{minipage}
        &
        \hspace{0.02\textwidth}
        \begin{minipage}{0.48\textwidth}
            \centering
            \includegraphics[width=\textwidth]{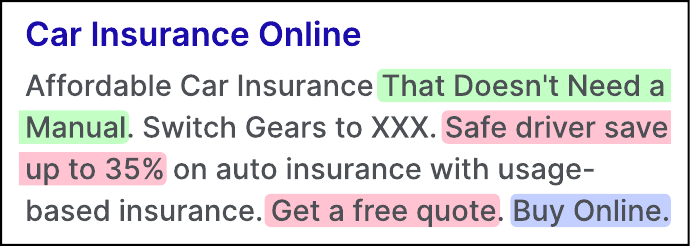}
            \caption{Example of \taskaaa~task. The highlighted text represents the \aaa~: \aaagreen{Features}, \aaared{Special deals}, and \aaablue{User-friendliness}.}
            \label{fig:example-a3}

            \quad

            \resizebox{\textwidth}{!}{%
                \begin{tabular}{@{}llc@{}}
                    \toprule
                    \multicolumn{2}{c}{\textbf{Sentences}} & \textbf{Score} \\ \midrule
                    \multirow{1}{*}{\textbf{S1}} & Suppon Black Vinegar with Luxury Ceramide & \multirow{2}{*}{5.00} \\
                    \multirow{1}{*}{\textbf{S2}} & Suppon Black Vinegar and Luxury Ceramide & \\
                    \midrule
                    \multirow{1}{*}{\textbf{S1}} & Find a gift that fits your budget & \multirow{2}{*}{2.33} \\
                    \multirow{1}{*}{\textbf{S2}} & Save up to 40\% on discounted products & \\
                    \bottomrule
                \end{tabular}%
            }
            \captionof{table}{Examples of \tasksim~task. S1 and S2 represent the paired Sentences 1 and 2, respectively.}
            \label{tab:example-sim}
        \end{minipage}
    \end{tabular}
\end{figure*}

\paragraph{\taskaccept.}
As most ad delivery platforms impose text length restrictions, minor grammatical errors are tolerated to enhance the readability and engage customers within limited space.
However, excessive compression can mislead customers, and such poor-quality ads should be detected before delivery to avoid negative impacts on the advertiser.
To assess this, we defined the \taskaccept~task, which predicts the acceptance of overall quality with binary labels: \texttt{acceptable}/\texttt{unacceptable}.
Based on expert feedback, unacceptable ad phenomena include ``collapsing symbols,'' ``unnatural repetition,'' ``incomprehensible meaning', and ``underspecified/vague''.
This differs from the general concept of linguistic acceptability, which checks for grammatical correctness, such as CoLA~\cite{warstadt2019cola}.
Examples of \taskaccept~are shown in Table \ref{tab:examples123}.
The ad text ``\emph{Engineer's career / Engineer's Jobs}'' is \emph{unacceptable} because the meaning is duplicated.
\paragraph{\taskconsist.}
Verifying consistency between the ad text and LP content is crucial.
If a feature or price mentioned in the ad text is not referenced in the corresponding LP, it may violate the Law for the Prevention of Unjustified Extra or Unexpected Benefit and Misleading Representation, resulting in damages to the advertiser.
However, these inconsistencies are difficult to detect as some factual expressions do not appear in LPs.
For example, the term ``\textit{official}'' is often used in ad text but rarely appears in LP content.
To assess this, we defined the \taskconsist~task, which predicts the consistency between LP content and ad text with binary labels: \texttt{consistent}/\texttt{inconsistent}.
Examples of \taskconsist~are shown in Table \ref{tab:examples123}.
The first line is labeled inconsistent because the LP refers to a ``\textit{single-family home},'' while the ad text mentions a ``\textit{condominium}.''

\paragraph{\taskpe.}
The most straightforward way to measure ad quality is to publish ads online and let end customers evaluate them.
However, delivering all ads without alterations is impractical, as low-quality ads can negatively impact advertisers.
Therefore, prior studies have investigated offline methods to measure ad text quality by simulating customer behavior, such as click-through rate (CTR),  based on past delivery history.
These methods are currently standard practices in many organizations.
Inspired by such works \cite{Gharibshah2020ctrpredictionlstm,niu2020ctrpredictioncnn,yang2022ctr}, we adopted the \taskpe~task to estimate a quality score in the range of [0, 100] based on ad texts, keywords, and industry, as shown in Table~\ref{tab:example-pe}.
The score simulates customer behavior based on past delivery history and is non-linearly transformed to maintain the original label distribution for contractual reasons.

\paragraph{\taskaaa.}
One of the most crucial factors in advertising is the \textit{aspect of advertising appeals} (\aaa).
At its core, advertising aims to connect advertisers with customers, and \aaa~serves as a bridge between them.
For example, an ad emphasizing \textit{low cost} may resonate with price-conscious customers, while one focusing on \textit{high performance} may not.
Thus, recognizing appealing expressions in advertising and using appropriate \aaa~can enhance downstream tasks, such as CTR prediction \cite{murakami2022aspect}.
In the \taskaaa~task, we follow a previous study \cite{murakami2022aspect}, which predicted all relevant \aaa~labels in a given ad text.
Figure \ref{fig:example-a3} shows an example of ad texts and the corresponding \aaa s.
All labels and distributions can be found in Table \ref{tab:label-distribution}.

\paragraph{\tasksim.}
Repeatedly showing the same ads to customers leads to \emph{ad fatigue}~\cite{abrams2007fatigue} and ad performance declines.
Therefore, it is essential to avoid displaying the same ads for extended periods and regularly replace them with different ones.
However, the transition from old to new ads must be carefully managed to maintain the product and its appeal.
Specifically, while the wording and representations are being updated, there is a risk of disengagement of customers who were attracted to previous ads.
Thus, when measuring similarity, we should particularly focus on this situation, which enables us to determine whether to replace the ad based on a quantified score.

Building on this motivation, we defined the \tasksim~task, which predicts the similarity score for an ad text pair on a scale of [1, 5].
The lower the values, the less similar the pair, and vice versa.
Examples are shown in Table \ref{tab:example-sim}.
The first example pair illustrates high similarity, where both the product of ``\textit{Suppon Black Vinegar}'' and the \aaa~of \textit{luxury} are identical.
Conversely, the second example pair differ in \aaa~with \textit{budget} and \textit{discount}, resulting in relatively low similarity.

\subsection{Dataset Construction}\label{sec:data-construction}

\begin{figure}[t]
    \centering
    \includegraphics[width=0.45\textwidth]{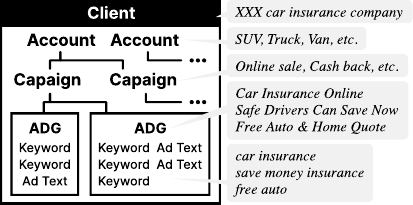}
    \caption{Account structure in ad delivery is hierarchical. A client represents a single company, and the account typically encompasses the commercial products offered by that client. Campaigns are created to promote these commercial products, while ad groups are used to organize keywords and ad texts. At higher levels of the hierarchy, there are more ads and greater variance. Conversely, at lower levels, there are fewer ads, which tend to be similar.}
    \label{fig:ad-structure}
\end{figure}

\begin{table*}[t]
\centering
\resizebox{\textwidth}{!}{%
    \begin{tabular}{@{}ccc@{}}
    \toprule
    \textbf{Task} & \textbf{Setup: Input} $\rightarrow$ \textbf{Label} & \textbf{Metrics} \\
    \midrule
    \multicolumn{1}{l}{\taskaccept}  & Classification: Ad text $\rightarrow$ \texttt{acceptable/unacceptable}      & Accuracy/F1-score \\
    \multicolumn{1}{l}{\taskconsist} & Classification: (Ad text, LP text) $\rightarrow$ \texttt{consistent/inconsistent}      & Accuracy/F1-score \\
    \multicolumn{1}{l}{\textsc{Ad Perf. Est.}} & Regression: (Ad texts, Keyword, Industry) $\rightarrow$ $[0, 100]$ & Pearson/Spearman corr. \\
    \multicolumn{1}{l}{\taskaaa}     & Classification: Ad text $\rightarrow$ multi-labels (See Appendix Table \ref{tab:label-distribution}) & F1-micro/-macro \\
    \multicolumn{1}{l}{\tasksim}     & Regression: Ad text pair $\rightarrow$ $[1, 5]$       & Pearson/Spearman corr. \\
    \bottomrule
    \end{tabular}%
}
\caption{Task descriptions of \mybenchmark. `\textsc{Ad Perf. Est.}' and `Corr.' are abbreviations for \taskpe~and correlation coefficient, respectively.}
\label{tab:task-overview}
\end{table*}

\paragraph{Data Collection.}
For the \textsc{Ad Accepta-bility}~and \textsc{Ad Consistency} tasks, data were collected from the ad creation phase of the actual AdOps workflow, including both human creators and NLG model outputs.
For the \taskpe~ and \tasksim~tasks, we used Japanese sponsored search ads delivered between 2021 and 2022.
For the \taskaaa~task, we used data from \citet{murakami2022aspect}.

\paragraph{Data Pre-processing.}
In the \taskpe~task, we negotiated with our clients to use a subset of the data approved for public release, as the data include ad performance metrics that are sensitive to advertisers.
Furthermore, we applied a nonlinear transformation to the raw CTR, scaling the values to the range of $[0,100]$ to preserve their distribution. Additionally, we masked proper nouns (e.g., product and company names) to prevent identification of advertisers and avoid any potential negative repercussions upon data release.
In the \tasksim~task, creating sentence pairs through random sampling proved inefficient because most pairs are not similar.
Therefore, we utilized the account structure configured during ad delivery, which includes \textit{client}, \textit{account}, \textit{campaign}, \textit{ad group}, and \textit{keyword} information, as depicted in Figure \ref{fig:ad-structure}. 
We created pseudo-similar pairs by sampling texts from the same \textit{ad group}, as preliminary results suggested.
To balance the label distribution, we created pseudo-dissimilar pairs by ensuring that the two texts belonged to different \textit{clients}.
Consequently, we sampled pseudo-similar and pseudo-dissimilar pairs at a ratio of 9:1.

\begin{table}[t]
\centering
\resizebox{0.40\textwidth}{!}{%
    \begin{tabular}{@{}lrcc@{}}
    \toprule
    \textbf{Task} & \multicolumn{1}{c}{\textbf{Train}} & \multicolumn{1}{c}{\textbf{Dev}} & \multicolumn{1}{c}{\textbf{Test}} \\
    \midrule
    \taskaccept  & 13,265  & 970 & 980  \\
    \taskconsist & 10,635  & 945 & 970  \\
    \textsc{Ad Perf. Est.} & 125,087 & 965 & 965 \\
    \taskaaa     & 1,856   & 465 & 410  \\
    \tasksim     & 4,980   & 623 & 629  \\
    \bottomrule
    \end{tabular}%
}
\caption{Number of instances for each task in each dataset.}
\label{tab:task-stats}
\end{table}

\paragraph{Annotation Workflow.}
All annotators were native Japanese speakers and experts with at least two years of professional experience in AdOps.
The annotation workflow for all tasks, except \taskpe~and \taskaaa, followed these steps:
(1) We first removed duplicate entries and filtered out data in languages other than Japanese.
(2) We iteratively revised the annotation guidelines until we achieved a satisfactory level of agreement through pilot annotations on small sampled datasets.
(3) Following the guidelines, we conducted a pilot annotation by asking three annotators to annotate the same sampled data used in Step 2.
Thereafter, we compared the annotators' results with our own and resolved any inconsistencies to further refine the guidelines.
This cycle was repeated at least twice.
(4) After completing Steps 1 to 3, we conducted the main annotation on the full test set using the finalized guidelines.
The complete guidelines are provided in Appendix \ref{sec:add-data}.

\paragraph{Data Splitting.}
Despite efforts at deduplication, similar ad expressions can still be easily found, potentially leading to data leakage with a simple random split.
Furthermore, assuming the data are used in industry, it is crucial to generalize effectively without overfitting to specific ad expressions.
Therefore, for \taskaccept~and \taskconsist, we split the data, considering the ad hierarchical structure in Figure \ref{fig:ad-structure}, ensuring that \textit{clients} do not overlap across training, development, and testing.
For \taskpe, we used the delivery structure, with the non-overlapping layer as the campaign.
For \taskaaa, we used the same split from \citet{murakami2022aspect}.
For the \tasksim, we randomly split the data to maintain label distribution consistency. 

\begin{table*}[t]
\centering
\resizebox{\textwidth}{!}{%
\begin{tabular}{@{}lccccc@{}}
\toprule
\multirow{2}{*}{\textbf{Evaluator}}
    & \textbf{\textsc{Ad Accept.}}
    & \textbf{\textsc{Ad Consist.}}
    & \textbf{\textsc{Ad Perf. Est.}}
    & \textbf{\taskaaa}
    & \textbf{\tasksim} \\
& \footnotesize{Accuracy/F1-score}
& \footnotesize{Accuracy/F1-score}
& \footnotesize{Pearson/Spearman}
& \footnotesize{F1-micro/-macro}
& \footnotesize{Pearson/Spearman} \\
\midrule
\multicolumn{6}{c}{\textit{Fine-tuned Encoder Models}} \\[0.1cm]
Tohoku BERT & \underline{0.711}/0.688 & \textbf{\underline{0.767}}/\underline{0.552} & \textbf{\underline{0.480}}*/\textbf{\underline{0.497}}* & 0.774/\textbf{\underline{0.694}}* & 0.773/0.807                   \\
Waseda BERT & 0.598/0.637 & 0.755/0.474          & 0.445/0.457                   & 0.663/0.517          & 0.740/0.800                   \\
XLM-RoBERTa & 0.705/\underline{0.690} & 0.758/0.519          & 0.453/0.457                   & \textbf{\underline{0.778}}/0.677 & \textbf{\underline{0.878}}*/\textbf{\underline{0.878}}* \\
\cmidrule{1-6}
\multicolumn{6}{c}{\textit{Zero-/Few-shot LLMs}} \\[0.1cm]
\calm       & \underline{0.520}/0.115 & 0.381/0.472          & 0.006/0.013                   & 0.154/0.042          & 0.036/0.036                   \\
\elyza      & 0.352/\underline{0.520} & \underline{0.628}/\underline{0.771}          & 0.003/0.046                   & 0.196/0.044          & 0.015/-0.004                  \\
GPT-4       & 0.325/0.433 & 0.583/0.612          & \underline{0.028}/\underline{0.073}                   & \underline{0.417}/\underline{0.113}          & \underline{0.776}/\underline{0.811}                   \\
\cmidrule{1-6}\morecmidrules\cmidrule{1-6}
Human & \textbf{0.732}*/\textbf{0.790}* & 0.703/\textbf{0.807}* & --- & 0.564/0.538 & 0.699/0.765 \\
\bottomrule
\end{tabular}%
}
\caption{Performance of PLMs and human evaluators on the test set. \underline{Underlined} indicates the best result for each setting, and \textbf{bold} indicates the best result across all methods. We conducted a t-test to compare the best result for each metric with the second-best, considering results statistically significant at p < 0.05. Statistically significant results are marked with an asterisk (*). See Appendix Table \ref{tab:eval-res-full} for the full results. 
}
\label{tab:eval-res}
\end{table*}

Table \ref{tab:task-stats} presents the statistics of our dataset.
The full tables of label distribution for each task are provided in Appendix Table \ref{tab:label-distribution}.

\section{Experimental Settings}

Table \ref{tab:task-overview} provides a brief overview of each task and the corresponding metrics used to measure task performance.
Our tasks are categorized into three setups: binary classification, multi-label classification, and regression.
For binary classification, we use accuracy and F1-score to evaluate the binary labels.
In multi-label classification, we follow \citet{murakami2022aspect} and use the F1-score with both macro and micro settings.
For regression, we use Pearson and Spearman correlation coefficients.

We employed two types of evaluators: the PLMs, including both fine-tuned settings with encoder models and zero-/few-shot settings for LLMs, as well as human evaluators.
Note that the detailed model description, hyperparameters, prompts, and additional information are provided in Appendix \ref{sec:exp-add}.

\paragraph{Fine-tuning Setting with Encoder Models.}
We utilized publicly available encoder models as baselines, specifically Tohoku BERT, Waseda RoBERTa, and XLM-RoBERTa~\cite{conneau2019xlmroberta}, which are commonly employed for Japanese NLP tasks.
These models differ in pre-tokenizer, tokenization unit, and pre-training dataset.
All the aforementioned models are of LARGE size.

\paragraph{Zero-/Few-shot Setting with LLMs.}
We employed \calm~and \elyza as baselines for open LLMs.
\calm~and \elyza~are based on the Llama 2~\cite{touvron2023llama2} architecture but differ in training methods and data; \calm~was trained from scratch, whereas \elyza~was continuously trained from the original Llama 2.\footnote{To further assess the gap between fine-tuning and zero-/few-shot learning of LLMs, we also fine-tuned an LLM. Due to the additional nature of this experiment, which falls beyond the scope of our study and is constrained by our computational resources, we selected only the best-performing open-source model from the zero-/few-shot settings. The settings and results are provided in Appendix \ref{sec:exp-add}.}

\paragraph{Human Evaluators.}
We enlisted three human evaluators who are \textit{not} involved in AdOps to evaluate all tasks except \taskpe.
We followed the same procedure for instruction as described in \S\ref{sec:data-construction}.
Pilot evaluations were carried out twice on 100 randomly sampled instances from the training set for each task before the main run.
In the final assessment, a majority vote per instance was conducted for the \taskaccept~and \taskconsist~tasks, while the average scores of evaluator assessments were reported for other tasks.

\section{Results and Discussion}

Table \ref{tab:eval-res} provides an overview of the results.
For the complete result including base size encoder models and GPT-3.5, please refer to Appendix Table \ref{tab:eval-res-full}.

\paragraph{Fine-tuned Setting with Encoder Models.~}
XLM-RoBERTa and Tohoku BERT achieved the highest or competitive scores in two or more tasks. 
In addition, the LARGE model outperformed the BASE model in most tasks, suggesting that an increased parameter size plays a key role in understanding ad expressions.

\paragraph{Zero-/Few-shot Setting with LLMs.}
GPT-4 achieved high performance across all three tasks,
while \elyza~performed the best among the LLMs in \taskconsist.
A substantial difference was observed between the open LLMs and OpenAI's LLMs in the \tasksim~task.
\calm~and \elyza~scored close to 0, indicating no correlation, whereas OpenAI's models, especially GPT-4, achieved competitive scores of 0.776/0.811.
Thus, the open LLMs used in this study struggle to handle semantic similarity or numerical answers, whereas GPT-4 performs considerably better.
In the \taskpe~task, all LLMs produced uncorrelated responses close to 0, indicating that accurately predicting ad performance remains a challenge for open LLMs.

\paragraph{Fine-tuned Encoder Models vs. Zero-/Few-shot LLMs.}
Overall, the fine-tuned models outperformed the LLMs.
The difference was substantial, ranging from 0.2 to 0.6, particularly for the \taskaccept, \taskpe, and \taskaaa~tasks.
This can be attributed to the characteristics of the tasks; \taskpe~involves predicting numbers in the range $[0,100]$, and \taskaaa~requires selecting all suitable labels from more than 20 labels, suggesting that the variety of data features and outputs could not be handled by few-shot alone.
Both \taskaccept~and \taskconsist~are binary classification tasks, yet the performance gap between fine-tuned models and LLMs is more noticeable in the \taskaccept~task, with a difference as large as 0.2 points, compared to the relatively smaller difference observed in the \taskconsist~task.
The task with the smallest difference was \tasksim, with a difference of only 0.06 to 0.10. 

\paragraph{PLMs vs. Humans.}
Human evaluators outper-form models in \taskaccept~and \taskconsist.
In both tasks, models tends to have high accuracy and low F1-score, while humans exhibit the opposite trend.
The evaluations by PLMs demonstrate high precision but low recall in both tasks, resulting in a low F1-score.
According to Table \ref{tab:label-distribution}, both datasets exhibit an imbalanced label distribution.
Models tend to predict the majority label, which likely leads to high accuracy.
In contrast, human experts achieve a balanced precision and recall, demonstrating robustness against the imbalanced label distribution and suggesting a higher level of generalization performance compared to the PLMs.
Particularly in the \taskconsist~task, where the label distribution is unbalanced, even the best model achieves an F1-score of only 0.55, whereas human performance reaches 0.8.
This suggests that humans can make better predictions in both precision and recall on unbalanced data.
Meanwhile, fine-tuned models outperform humans on the \taskaaa~and \tasksim~tasks.
In the \taskaaa~task, human evaluators struggled with the diversity of output labels, similar to LLMs.
However, the difference between the F1-micro and F1-macro scores is relatively small at 0.03 points for humans.
In contrast, for the fine-tuned PLMs, the difference ranges from 0.08 to 0.20 points.
This indicates that human evaluators have a strong ability to generalize and can maintain higher performance, even when a label appears infrequently.
In the \tasksim~task, GPT-4, in addition to fine-tuned models, outperforms human evaluators.
This is the only task where LLM outperforms humans, suggesting that GPT-4 has a high level of alignment with expert humans in terms of semantic similarity and numerical understanding.

\begin{table}[t]
    \centering
    \resizebox{0.43\textwidth}{!}{%
        \begin{tabular}{@{}ccccrr@{}}
            \midrule
            \textbf{Error} & \textbf{GT} & \textbf{H} & \textbf{M} & \textbf{\textsc{Accept.}} & \textbf{\textsc{Consist.}}\\
            \midrule
            \multirow{2}{*}{\textbf{C-IC}}& \texttt{T} & \texttt{T} & \texttt{F} &
                \cellcolor[rgb]{0.341, 0.733, 0.541}30.1\% & \cellcolor[rgb]{0.631, 0.851, 0.745}17.2\% \\
            & \texttt{F} & \texttt{F} & \texttt{T} &
                \cellcolor[rgb]{0.999, 0.999, 0.999}0.6\% & \cellcolor[rgb]{0.894, 0.957, 0.925}5.5\% \\
            \cmidrule{1-6}
            \multirow{2}{*}{\textbf{IC-C}}& \texttt{T} & \texttt{F} & \texttt{T} &
                \cellcolor[rgb]{0.999, 0.999, 0.999}0.6\% & \cellcolor[rgb]{0.973, 0.992, 0.984}1.9\% \\
            & \texttt{F} & \texttt{T} & \texttt{F} &
                \cellcolor[rgb]{0.612, 0.843, 0.729}18.0\% & \cellcolor[rgb]{0.773, 0.910, 0.843}10.9\% \\
            \cmidrule{1-6}
            \multirow{2}{*}{\textbf{IC-IC}}& \texttt{T} & \texttt{F} & \texttt{F} &
                \cellcolor[rgb]{0.900, 0.965, 0.937}4.9\% & \cellcolor[rgb]{0.878, 0.953, 0.914}6.2\% \\
            & \texttt{F} & \texttt{T} & \texttt{T} &
                \cellcolor[rgb]{0.929, 0.973, 0.953}3.9\% & \cellcolor[rgb]{0.871, 0.949, 0.910}6.6\%  \\
            \bottomrule
        \end{tabular}
    }
    \caption{Type-specific error rates for Tohoku BERT (BERT) and XLM-RoBERTa (XLM-R) in the \taskaccept~and \taskconsist~task. Ground truth (GT), human (H), and model (M) labels are represented as T (\texttt{acceptable} or \texttt{consistent}) and F (\texttt{unacceptable} or \texttt{inconsistent}). }%
    \label{tab:human-model-confmtx-b}
\end{table}
\begin{figure*}[t]
    \centering
    \includegraphics[width=0.95\textwidth]{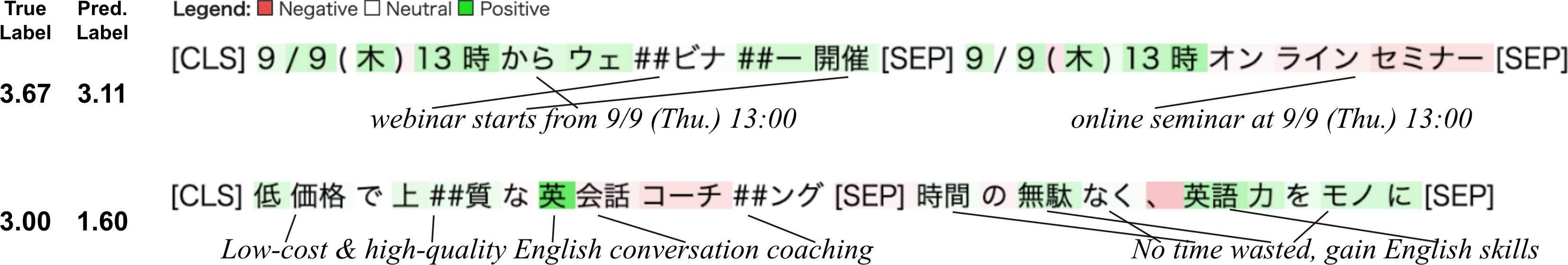}
    \caption{Examples of integrated gradient visualization with Tohoku BERT model's outputs showing the difference in attention for small (top) and large (bottom) gaps between ground truth and predicted labels in the \tasksim~task. Red indicates a negative influence, while green indicates positive influence on the predictions.}
    \label{fig:ig-adsim}
\end{figure*}

\paragraph{Error Analysis for \taskaccept~and \taskconsist~Tasks.}
We conducted a detailed analysis of two tasks: \taskaccept~and \taskconsist, where the model has not yet outperformed human evaluators.
We analyzed three types of errors: human incorrect and model correct (IC-C), human correct and model incorrect (C-IC), and both incorrect (IC-IC), as shown in Table \ref{tab:human-model-confmtx-b}.
In both tasks, the model more frequently predicted False labels than True labels for both IC-C and C-IC errors.
In contrast, humans provided more True labels in both correct and incorrect cases.
This suggests that humans exhibit a relatively higher degree of leniency compared to models, which tend to be overly cautious when making decisions in the \taskaccept~and \taskconsist~tasks.

Additionally, in the \taskaccept~task, the model can easily identify elements that can be detected heuristically, such as ``incorrect use of symbols'' or grammatically incorrect or non-fluent expressions.
However, it struggles with evaluating expressions that are fluent but semantically unnatural, such as ``\textit{Engineer’s career / Engineer’s Job}'' (redundancy in the same appeal) or ``\textit{Condominium Sales / Single-family Home}'' (contradictory appeals).
To address these challenges, strategies such as adding training data or providing the LLM with more detailed judgment criteria could enhance accuracy.

Regarding the \taskconsist~task, the model generally performs well when expressions from the LP are included in the ad text.
However, it tends to make errors when this is not the case. There are two potential causes for this scenario.
One is the use of meta-expressions, such as the word ``\textit{official}.''
In ad text, ``\textit{official}'' is often used to reassure customers, but it is frequently omitted in LPs because there is no need to emphasize it.
To address this issue, accuracy might be improved by further training on similar data or by incorporating a specific mechanism to infer whether the LP is ``\textit{official}.''
The second cause involves referring to non-verbal information, such as images, in the ad text.
Although this is beyond the scope of the current study, it might be addressed by leveraging multimodal models.

\paragraph{Case Study on the Model's Behavior on \tasksim~Task.}
We used integrated gradients~\cite{sundararajan2017ig} to visualize the attention of Tohoku BERT, aiming for a deeper understanding of the model's behavior, which can potentially enhance its performance.
Figure \ref{fig:ig-adsim} shows examples of low and high errors between predictions and the ground truth in the \tasksim~task.
In the first example, the model fails to pay sufficient attention to ``\textit{online seminar},'' which has a similar meaning to ``\textit{webinar}.''
However, high attention to the date expression ``\textit{9/9 (Thursday) at 13:00}'' allows the model to predict a score close to the correct answer.
However, in the second example, the model did not pay enough attention to ``\textit{English conversation coaching},'' which has a similar meaning to ``\textit{gaining English skills},'' resulting in an incorrect prediction of dissimilarity.
As we found many such cases, we believe that the model tends to prioritize surface-level information, especially for entities such as dates, times, and numbers.
This may prevent the model from correctly identifying cases that are semantically similar, even when the surface-level information differs, and vice versa.

Another instance where we observed shortcomings in the models is in reasoning.
For example, the models mistakenly interpret the phrases ``\textit{half price}'' and ``\textit{buy 2 get 1 free}'' as having the same meaning, whereas humans easily comprehend the difference.
Additionally, pairs of phrases like ``\textit{Available 24 hours a day, 7 days a week}'' and ``\textit{Closed every Wednesday},'' as well as ``\textit{Property located near the station}'' and ``\textit{20-minute walk from the station}'' often confuse the models in understanding.
These types of paraphrasing are commonly used in the ad creation process; however, it is challenging to automatically capture them in a natural dataset.
Therefore, a specialized dataset focusing on these phenomena is necessary to enable models to capture them accurately.


\section{Related Work}


\paragraph{Evaluating Ad Text Quality}
\citet{mita-etal-2024-striking}~proposed an ad text generation benchmark that aims to evaluate NLG models (Step 2 in Figure \ref{fig:adops}) with surface- and semantic-level overlap metrics, such as BLEU~\cite{papineni-etal-2002-bleu}, ROUGE~\cite{lin-2004-rouge}, and BERTScore~\cite{bertscore}, along with simple heuristics, including keyword insertion rate~\cite{Mishra-et-al:20} and text length.
However, this approach is inadequate for real-world applications because (1) it requires reference texts to assess generated texts, making it difficult to evaluate new ads that have no references, and (2) simple heuristics cannot guarantee the quality of the ad text for delivery, as more specialized criteria are required in actual production, such as appropriate wording, effective appeals, consistency between ad text and product information, and high predicted performance.

Ad text quality is typically represented by metrics of user behavior obtained after ad delivery as illustrated in Step 6 of Figure \ref{fig:adops}, and CTR is widely used to predict ad performance and make marketing strategies.
The CTR prediction task was first studied by \citet{robinson2007adeffect} and has been actively researched by \citet{rosales2012ctr}, \citet{mcmahan2013ctr}, \citet{Chapelle2015ctr}, \citet{yan2014ctr}, and \citet{kumar2015ctr}, alongside traditional machine learning techniques, such as logistic regression-based and factorization machine (FM)-based models~\cite{rendle2010fm} over the past decades~\cite{yang2022ctr}.
While there are many methods for CTR prediction, few studies have evaluated the quality of ad texts at a more granular level.
Coarse-level feedback may be insufficient to improve ad quality, as it is unclear where edits should be made or what changes are necessary.
Meanwhile, \mybenchmark~provides more detailed evaluation results, such as acceptability and consistency, making it easier to integrate into actual AdOps workflows.

\section{Conclusion}
We defined five tasks to verify the quality of ad texts and developed the \mybenchmark, a large, versatile, and comprehensive benchmark of Japanese advertising data constructed for the NLP community, based on real-world AdOps workflows.
We conducted evaluations with both PLMs and human evaluators on \mybenchmark~to explore its characteristics, offering insights into practical workflow applications and identifying potential areas for future improvement and research.
Our findings suggest that sampling directly from real data generally benefits the model, highlighting the importance of tasks centered on natural language inference and semantic understanding.
We hope that the combination of our defined tasks and datasets will advance research in ad text evaluation, bridging the fields of advertising and NLP and paving the way for new discoveries and applications.

\section*{Limitations}


\paragraph{Language.}
The language was limited to Japanese, as the advertising agency is located in Japan.
Expressions related to online advertising vary according to language and culture.
For instance, in Japan, for example, there is a cultural preference for cluttered design~\cite{takahiko2001japanesedesign}, which results in a lot of information being scattered around, compared to the clean LPs in the US and Europe.
This means that different strategies and models may be required for different languages and countries. 
Although exploring the adaptability of datasets across languages is beyond the scope of this study, it represents an exciting opportunity for future research, especially because the field of NLP in advertising remains largely unexplored.

Additionally, existing research indicates that machine translation techniques could potentially mitigate these limitations \cite{miyazaki-shimizu-2016-cross,masad-etal-2023-automatic,wang-hershcovich-2023-evaluating}.
Moreover, the methodology for designing these tasks and datasets is language-independent, making it applicable to other languages. Tasks such as assessing the quality of ad text (\taskaccept), checking the consistency of ad text with landing pages (\taskconsist), and rephrasing ad texts (\tasksim) can enhance operational efficiency across different languages. By referencing the motivations and methods outlined in this paper, future researchers and engineers can design similar tasks without starting from scratch when working in different languages.

Finally, it is important to note that the primary contribution of our study is the definition of real-world advertising operations as distinct tasks and the integration of these tasks into a comprehensive benchmark, using the Japanese dataset as a case study.

\paragraph{Image Data.}
This study focuses solely on textual data, with image data excluded from the scope for the following reasons: (1) search engine ads consist only of text data, (2) LP images often convey content through text, (3) the performance improvement gained is small compared to the effort required to handle image data. Thus, we consider image data as supplementary in this study. While certain types of ads, such as display ads, primarily use images, further exploration of this area would be a promising direction for future work.

\paragraph{Data Bias.}
Since the clients are limited to those who have contracts with the advertising agency, the dataset may not entirely represent all industries.
Consequently, the findings of this study may not be fully generalizable to the broader internet advertising market.
Future research could benefit from including a wider range of data sources to address these limitations.

\section*{Ethical Consideration}
Our AdTEC dataset comprises documents such as ad texts, LP texts, keywords, and industry types. The dataset includes only a subset of the original data, sampled over a specific period and refined through preprocessing, as detailed in \S \ref{sec:data-construction} of our paper.

Each instance in the dataset consists of ad text data along with related information, and each instance is labeled, as discussed in the task design section (\S\ref{sec:task-design}). While the dataset is self-contained, it does not include any confidential information, as all data intended for internal use were made public through appropriate channels. However, it is important to note that the use of automatically generated ad texts by NLG models may introduce unnatural expressions, which could be considered a source of noise.

The data were collected in 2023 using both manual human curation and software programs. Authors and other full-time employees were involved in the data collection process.
The were compensated at rates above the minimum wage and provided with basic social security benefits. Although no formal ethical review was conducted, individuals involved were directly notified and their consent was obtained before data collection..

Preprocessing, cleaning, and labeling were performed on the data, and the raw data were saved for potential future uses. This dataset has already been used to evaluate ad text quality at the author's company and could be utilized for tasks such as language modeling on acceptable ad texts and hallucination detection. The dataset will be distributed to third parties via a URL and will be available under the Creative Commons Attribution-NonCommercial-ShareAlike 4.0 (CC-BY-NC-SA 4.0) International License. The authors will support and maintain the dataset, and updates may be made to correct errors or delete instances upon request. Users will be informed of any updates through our website or codebase.

Please refer to the Datasetsheets~\cite{gebru2021datasheets}, a framework designed to thoroughly cover essential information about the dataset, in Appendix \ref{sec:datadoc} for comprehensive and detailed information.

\section*{Acknowledgements}
We would like to thank Yoshiko Shinozuka and other members of the KiwamiTD\footnote{\texttt{\url{https://www.cyberagent.co.jp/news/detail/id=24670}}} team at CyberAgent for their valuable comments and assistance with the background research and data collection.

\bibliography{custom}

\appendix

\clearpage


\section{Datasheets}\label{sec:datadoc}
{We adhere to the existing dataset documentation framework proposed by \citet{gebru2021datasheets} to provide comprehensive information about our dataset.

\subsection{Motivation}

\paragraph{For what purpose was the dataset created?}
This dataset is designed to evaluate the quality of ad texts in multiple aspects from the perspective of practical advertising operations.

\paragraph{Who created the dataset (for example, which team, research group) and on behalf of which entity (for example, company, institution, organization)?}
The authors of this paper created the dataset and conducted the research.

\paragraph{Who funded the creation of the dataset?}
N/A

\subsection{Composition}

\paragraph{What do the instances that comprise the dataset represent (for example, documents, photos, people, countries)?}
The dataset represents documents, e.g., ad texts, LP texts, keywords, and industry types.

\paragraph{How many instances are there in total (of each type, if appropriate)?}
See Table \ref{tab:task-stats} and \ref{tab:task-overview}.

\paragraph{Does the dataset contain all possible instances or is it a sample (not necessarily random) of instances from a larger set?}
As described in \ref{sec:data-construction}, the dataset includes only a subset of the original data, sampled over a specific period and refined through preprocessing.

\paragraph{What data does each instance consist of?}
Our dataset primarily comprises ad text data and related information, including LP texts, keywords, and industry types.

\paragraph{Is there a label or target associated with each instance?}
Yes. See Section \ref{sec:task-design}.

\paragraph{Is any information missing from individual instances?}
N/A.

\paragraph{Are relationships between individual instances made explicit (for example, users’ movie ratings, social network links)?}
N/A.

\paragraph{Are there recommended data splits (for example, training, development/validation, testing)?}
Yes. See Section \ref{sec:data-construction} and Table \ref{tab:task-stats}.

\paragraph{Are there any errors, sources of noise, or redundancies in the dataset?}
The use of automatically generated ad texts by NLG models may introduce unnatural expressions, which can be considered a form of noise.

\paragraph{Is the dataset self-contained, or does it link to or otherwise rely on external resources (for example, websites, tweets, other datasets)?}
The dataset is self-contained for the tasks described in the paper.

\paragraph{Does the dataset contain data that might be considered confidential (for example, data that is protected by legal privilege or by doctor–patient confidentiality, data that includes the content of individuals’ non-public communications)?}
Data originally intended for internal use but subsequently made public through appropriate channels.

\paragraph{Does the dataset contain data that, if viewed directly, might be offensive, insulting, threatening, or might otherwise cause anxiety?}
No.

\paragraph{Does the dataset identify any subpopulations (for example, by age, gender)?}
No.

\paragraph{Is it possible to identify individuals (that is, one or more natural persons), either directly or indirectly (that is, in combination with other data) from the dataset?}
No.

\paragraph{Does the dataset contain data that might be considered sensitive in any way (for example, data that reveals race or ethnic origins, sexual orientations, religious beliefs, political opinions or union memberships, or locations; financial or health data; biometric or genetic data; forms of government identification, such as social security numbers; criminal history)?}
No.

\subsection{Collection Process}

\paragraph{How was the data associated with each instance acquired? Was the data directly observable (for example, raw text, movie ratings), reported by subjects (for example, survey responses), or indirectly inferred/ derived from other data (for example, part-of-speech tags, model-based guesses for age or language)?}
See Section \ref{sec:data-construction}.

\paragraph{What mechanisms or procedures were used to collect the data (for example, hardware apparatuses or sensors, manual human curation, software programs, software APIs)?}
See Section \ref{sec:data-construction}.

\paragraph{If the dataset is a sample from a larger set, what was the sampling strategy (for example, deterministic, probabilistic with specific sampling probabilities)?}
As described in Section \ref{sec:data-construction}, we utilized data from a specific period without employing probabilistic sampling. Items that could not be technically included in the dataset were removed.

\paragraph{Who was involved in the data collection process (for example, students, crowdworkers, contractors) and how were they compensated (for example, how much were crowdworkers paid)?}
Authors and other full-time employees. All workers involved in this research are employed as full-time employees by the company and are compensated at rates above the minimum wage set by local authorities. Additionally, they are provided with basic social security benefits.

\paragraph{Over what timeframe was the data collected?}
The data collection was conducted in 2023.

\paragraph{Were any ethical review processes conducted (for example, by an institutional review board)?}
No.

\paragraph{Did you collect the data from the individuals in question directly, or obtain it via third parties or other sources (for example, websites)?}
In question directly.

\paragraph{Were the individuals in question notified about the data collection?}
Yes.

\paragraph{Did the individuals in question consent to the collection and use of their data?}
Yes.

\paragraph{If consent was obtained, were the consenting individuals provided with a mechanism to revoke their consent in the future or for certain uses?}
N/A.

\paragraph{Has an analysis of the potential impact of the dataset and its use on data subjects (for example, a data protection impact analysis) been conducted?}
N/A.

\subsection{Preprocessing/cleaning/labeling}

\paragraph{Was any preprocessing/cleaning/labeling of the data done (for example, discretization or bucketing, tokenization, part-of-speech tagging, SIFT feature extraction, removal of instances, processing of missing values)?}
Yes. See Section \ref{sec:data-construction}.

\paragraph{Was the ``raw'' data saved in addition to the preprocessed/cleaned/ labeled data (for example, to support unanticipated future uses)?}
Yes.

\paragraph{Is the software that was used to preprocess/clean/label the data available?}
No.

\subsection{Uses}

\paragraph{Has the dataset been used for any tasks already?}
Yes. The dataset has been used to evaluate the quality of ad texts at the author's company. 

\paragraph{Is there a repository that links to any or all papers or systems that use the dataset?}
Yes. \texttt{\url{https://cyberagentailab.github.io/AdTEC}}

\paragraph{What (other) tasks could the dataset be used for?}
Various tasks using the dataset can be considered, such as language modeling on the acceptable ad texts and hallucination detection with the \taskconsist~dataset.

\paragraph{Is there anything about the composition of the dataset or the way it was collected and preprocessed/cleaned/labeled that might impact future uses?}
As mentioned in Section 4, the \taskpe~task masks certain entities to protect information. Consequently, users may need to complete the task with incomplete information, which could affect its usability.

\paragraph{Are there tasks for which the dataset should not be used?}
N/A.

\begin{table*}[t]
    \centering
    \resizebox{0.8\textwidth}{!}{%
        \begin{tabular}{ll}
            \toprule
            Name & URL \\
            \midrule
            Tohoku BERT & \href{https://huggingface.co/tohoku-nlp/bert-base-japanese-v2}{\texttt{tohoku-nlp/bert-\{base,large\}-japanese-v2}}\\
            Waseda RoBERTa & \href{https://huggingface.co/nlp-waseda/roberta-base-japanese-with-auto-jumanpp}{\texttt{nlp-waseda/roberta-\{base,large\}-japanese-with-auto-jumanpp}}\\
            XLM-RoBERTa & \href{https://huggingface.co/xlm-roberta-base}{\texttt{xlm-roberta-\{base,large\}}} \\
            CALM2 & \href{https://huggingface.co/cyberagent/calm2-7b-chat}{\texttt{cyberagent/calm2-7b-chat}}\\
            ELYZA & \href{https://huggingface.co/elyza/ELYZA-japanese-Llama-2-7b-instruct}{\texttt{elyza/ELYZA-japanese-Llama-2-7b-instruct}}\\
            OpenAI & \texttt{2023-08-01-preview} \\
            \bottomrule
        \end{tabular}
    }
    \caption{URLs and names of used PLMs.}
    \label{tab:model-names}
\end{table*}

\subsection{Distribution}

\paragraph{Will the dataset be distributed to third parties outside of the entity (for example, company, institution, organization) on behalf of which the dataset was created?}
Yes.

\paragraph{How will the dataset be distributed (for example, tarball on website, API, GitHub)?}
\texttt{\url{https://cyberagentailab.github.io/AdTEC}}.

\paragraph{When will the dataset be distributed?}
After this paper is accepted, as soon as possible.

\paragraph{Will the dataset be distributed under a copyright or other intellectual property (IP) license, and/or under applicable terms of use (ToU)?}
The dataset is distributed under Creative Commons Attribution-NonCommercial-ShareAlike 4.0 International (CC BY-NC-SA 4.0) License.

\paragraph{Have any third parties imposed IP-based or other restrictions on the data associated with the instances?}
The third party data is not intended for commercial use and is subject to the organization's terms and conditions.

\paragraph{Do any export controls or other regulatory restrictions apply to the dataset or to individual instances?}
No

\subsection{Maintenance}

\paragraph{Who will be supporting/hosting/maintaining the dataset?}
The authors will be.

\paragraph{How can the owner/curator/ manager of the dataset be contacted (for example, email address)?}
By the email address.

\paragraph{Is there an erratum?}
N/A.

\paragraph{Will the dataset be updated (for example, to correct labeling errors, add new instances, delete instances)?}
It is possible to update the dataset on the website or codebase to correct errors or delete instances upon request.

\paragraph{If the dataset relates to people, are there applicable limits on the retention of the data associated with the instances (for example, were the individuals in question told that their data would be retained for a fixed period of time and then deleted)?}
N/A.

\paragraph{Will older versions of the dataset continue to be supported/hosted/ maintained?}
It depends on the nature of the dataset update. We may inform dataset users through our website or codebase.

\paragraph{If others want to extend/augment/build on/contribute to the dataset, is there a mechanism for them to do so?}
No. We consider that the dataset should be used exclusively for our task. Any further work should be a separate contribution from ours.

\begin{table*}[t]
\centering
\resizebox{\textwidth}{!}{%
    \begin{tabular}{lcccccccc}
    \toprule
    \multicolumn{1}{l}{\textbf{Model}} & \multicolumn{1}{c}{\textbf{PreTokenizer (Dictionary)}} &\multicolumn{1}{c}{\textbf{Tokenization Unit}} & \multicolumn{1}{c}{\textbf{Dataset}} & \textbf{\#Vocab} & \textbf{\#Param} \\
    \midrule
    Tohoku BERT$_{\rm BASE}$ & MeCab (IPADic+NEologd) & BPE & Wikipedia (Ja) & 32K & 111M \\
    Tohoku BERT$_{\rm LARGE}$ & MeCab (IPADic+NEologd) & BPE & Wikipedia (Ja) & 32K & 337M\\
    Waseda RoBERTa$_{\rm BASE}$ & Juman++ & Unigram LM & Wikipedia (Ja) + CC (Ja) & 32K & 110M\\
    Waseda RoBERTa$_{\rm LARGE}$ & Juman++ & Unigram LM & Wikipedia (Ja) + CC (Ja) & 32K & 336M\\
    XLM-RoBERTa$_{\rm BASE}$ & --- & Unigram LM & Multilingual CC & 250K & 278M\\
    XLM-RoBERTa$_{\rm LARGE}$ & --- & Unigram LM & Multilingual CC & 250K & 559M\\
    \bottomrule
    \end{tabular}
}
\caption{List of trained language models used in the experiment, where CC and Ja represent CommonCrawl and Japanese data, respectively.}
\label{tab:model-detail}
\end{table*}
\begin{table}[t]
    \centering
    \resizebox{0.45\textwidth}{!}{%
        \begin{tabular}{lr}
        \toprule
        \textbf{Parameter} & \textbf{Value} \\
        \midrule
        \multicolumn{2}{c}{\textit{Fine-tuned Encoder Models}} \\[0.1cm]
        Learning Rate & \{2e-5, 5.5e-5, 2e-6\}   \\
        Seed & 0 \\
        Epoch & 30  \\
        Early Stopping Patience & 3  \\
        Optimizer & Adam  \\
        Max Sequence Length & 128 \\
        \cmidrule{1-2}
        \multicolumn{2}{c}{\textit{Zero-/Few-shot LLMs}} \\[0.1cm]
        Attempts Per an Instance & 5 \\
        Temperature & 0.8 \\
        Max New Tokens & 64 \\
        \cmidrule{1-2}
        \multicolumn{2}{c}{\textit{Fine-tuned LLMs}} \\[0.1cm]
        LoRA alpha & 16\\
        LoRA dropout & 0.1\\
        Bottleneck r & 16\\
        \bottomrule
        \end{tabular}%
    }
    \captionof{table}{Hyperparameters used to the fine-tuned encoder and LLM evaluators. Numbers in curly brackets represent the range of possible values.}
    \label{tab:hyperparam}
\end{table}

\section{Additional Information in Dataset Construction}\label{sec:add-data}


\begin{figure}[t]
    \centering
    \includegraphics[width=0.5\textwidth]{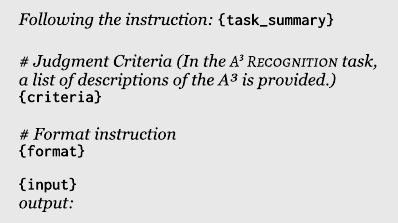}
    \caption{The prompt template for the LLM experiment. The text in italics and curly brackets represents the prompt text and the placeholders, respectively. The number of inputs corresponds to $n$ in the $n$-shot settings.}
    \label{fig:llm-prompt-template}
\end{figure}

\subsection{Annotation Guidelines}\label{sec:annotation-guideline}
The annotation guidelines for \taskaccept, \taskconsist, and \tasksim~are shown in Figures \ref{fig:anno-guide-1} and \ref{fig:anno-guide-2}.

\subsection{Label Distributions}
The label distributions for \taskaccept, \taskconsist, \taskaaa, \tasksim, and \taskpe~tasks are shown in Table \ref{tab:label-distribution} and Figure \ref{fig:label-distribution-pe}.

\begin{figure*}[t]
    \centering
    \begin{minipage}{0.42\textwidth}
        \resizebox{\textwidth}{!}{%
            \begin{tabular}{@{}lrrrr@{}}
            \toprule
            \multicolumn{1}{c}{\textbf{Label}} & \multicolumn{1}{c}{\textbf{Train}} & \multicolumn{1}{c}{\textbf{Dev}} & \multicolumn{1}{c}{\textbf{Test}} & \multicolumn{1}{c}{\textbf{Total}} \\
            \midrule
            \multicolumn{5}{c}{\taskaccept} \\[0.1cm]
            Acceptable     & 15,099 & \multicolumn{1}{c}{--} & 850   & 15,949 \\
            Not acceptable & 20,278 & \multicolumn{1}{c}{--} & 1,150 & 21,428 \\
            Total          & 35,377 & \multicolumn{1}{c}{--} & 2,000 & 37,377 \\
            \cmidrule{1-5}
            \multicolumn{5}{c}{\taskconsist} \\[0.1cm]
            Consistent     & 8,708  & \multicolumn{1}{c}{--} & 620   & 9,328  \\
            Not consistent & 20,048 & \multicolumn{1}{c}{--} & 1,380 & 21,428 \\
            Total          & 28,756 & \multicolumn{1}{c}{--} & 2,000 & 30,756 \\
            \cmidrule{1-5}
            \multicolumn{5}{c}{\taskaaa} \\[0.1cm]
            Special deals & 230 & 63 & 42 & 335 \\
            Discount price & 80 & 22 & 16 & 118 \\
            Reward points & 50 & 17 & 14 & 81 \\
            Free & 283 & 66 & 62 & 411 \\
            Special gift & 83 & 20 & 18 & 121 \\
            Features & 746 & 178 & 190 & 1,114 \\
            Quality & 35 & 17 & 11 & 63 \\
            Problem solving & 10 & 4 & 3 & 17 \\
            Speed & 95 & 23 & 20 & 138 \\
            User-friendliness & 213 & 65 & 32 & 310 \\
            Transportation & 53 & 17 & 10 & 80 \\
            Limited offers & 36 & 10 & 6 & 52 \\
            Limited time & 43 & 12 & 5 & 60 \\
            Limited target & 81 & 10 & 18 & 109 \\
            First-time limited & 16 & 5 & 4 & 25 \\
            Performance & 47 & 13 & 9 & 69 \\
            Largest/No.1 & 108 & 11 & 20 & 139 \\
            Product lineup & 167 & 38 & 42 & 247 \\
            Trend & 69 & 15 & 13 & 97 \\
            Others & 117 & 33 & 23 & 173 \\
            Story & 73 & 16 & 9 & 98 \\
            Total & 2,635 & 655 & 567 & 3,857 \\
            \cmidrule{1-5}
            \multicolumn{5}{c}{\taskaaa~(\#Labels per document)} \\[0.1cm]
            0 & 337 & 94 & 84 & 515 \\
            1 & 769 & 198 & 165 & 1,132 \\
            2 & 485 & 98 & 100 & 683 \\
            3 & 182 & 44 & 47 & 273 \\
            4 & 69 & 26 & 10 & 105 \\
            5 & 10 & 5 & 3 & 18 \\
            6 & 4 & 0 & 1 & 5 \\ 
            Total & 1,856 & 465 & 410 & 2,731 \\
            \cmidrule{1-5}
            \multicolumn{5}{c}{\tasksim} \\[0.1cm] 
            $1 \leq x < 2$ & 527 & 66 & 67 & 660  \\
            $2 \leq x < 3$ & 845 & 105 & 108 & 1,058 \\
            $3 \leq x < 4$ & 2,739 & 343 & 344 & 3,426 \\
            $4 \leq x < 5$ & 790 & 99 & 100 & 989  \\
            $5 \leq x$ & 79 & 10 & 10 & 99  \\ 
            Total & 4,980 & 623 & 629 & 6,232  \\
            \bottomrule
            \end{tabular}%
        }
        \captionof{table}{Label distribution of \taskaccept, \taskconsist, \taskaaa~and \tasksim task. Note that the number of labels in \taskaaa~does not necessarily equal the number of sentences, because a single document can have multiple labels.}
        \label{tab:label-distribution}
    \end{minipage}
    \hspace{0.5cm}
    \begin{minipage}{0.5\textwidth}
        \begin{minipage}[t]{\linewidth}
            \includegraphics[width=\textwidth]{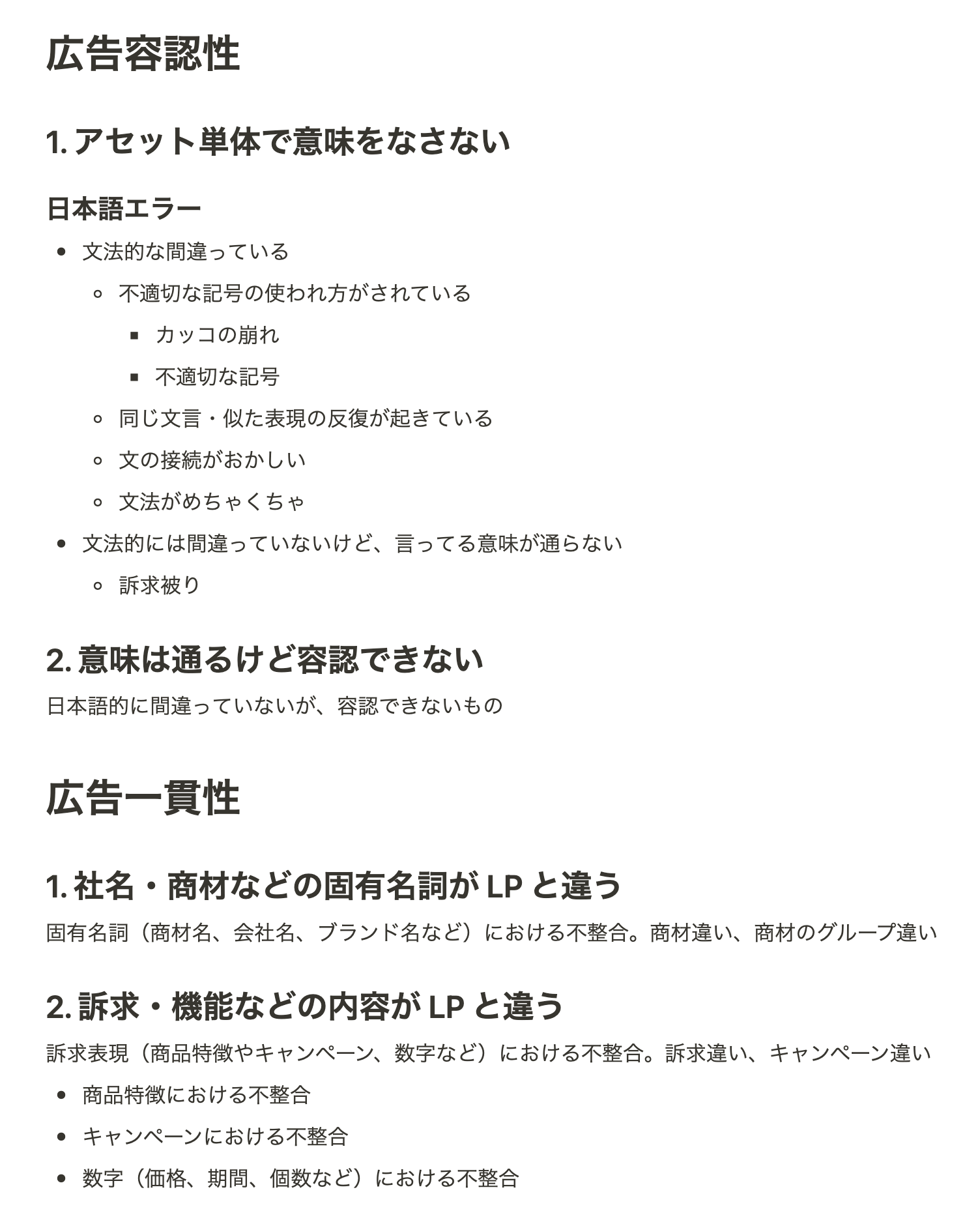}
            \caption{Annotation guideline for \taskaccept and \taskconsist~task.}
            \label{fig:anno-guide-1}
        \end{minipage}
        \begin{minipage}[t]{\linewidth}
            \includegraphics[width=\textwidth]{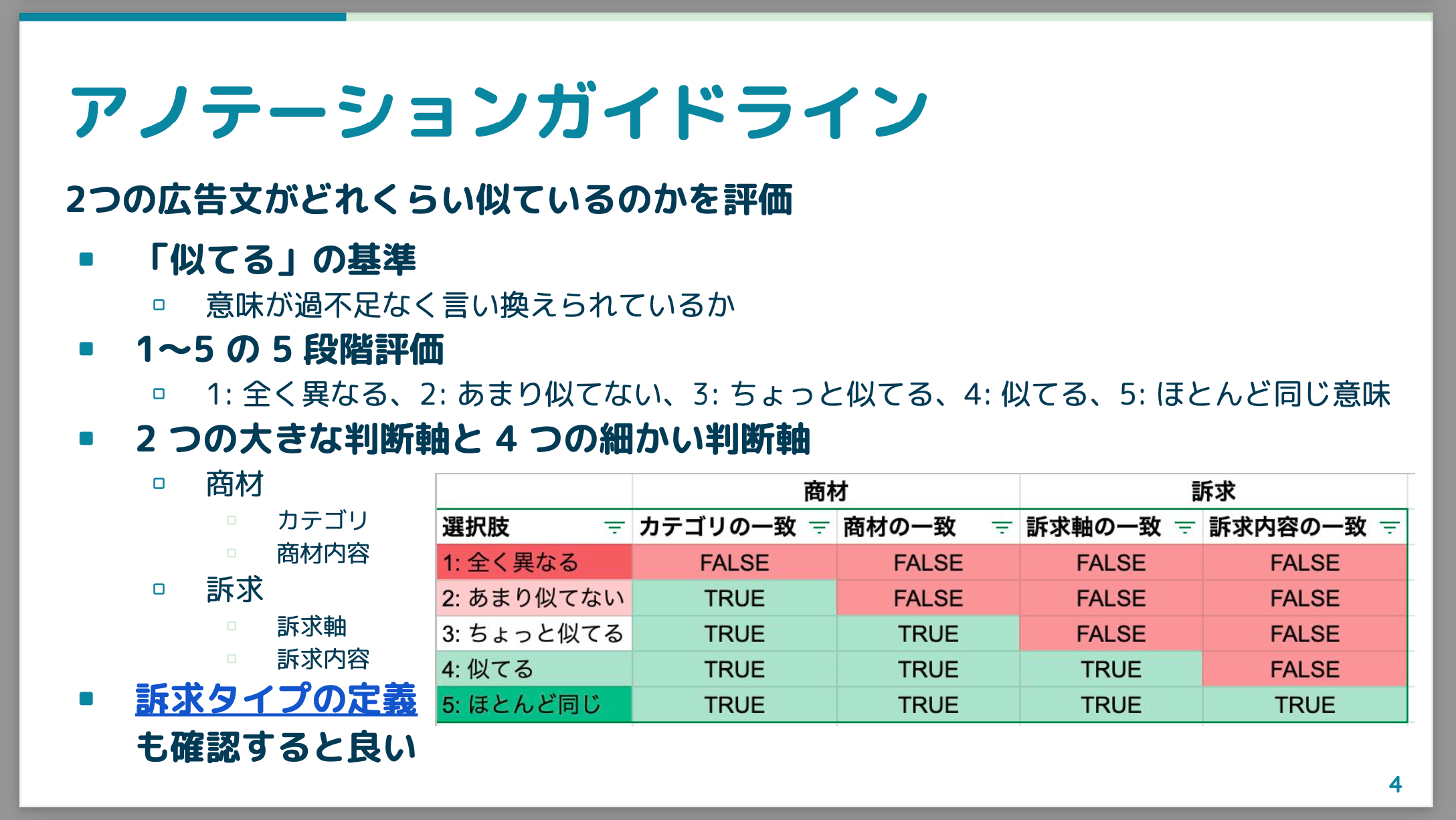}
            \caption{Annotation guideline for \tasksim~task.}
            \label{fig:anno-guide-2}
        \end{minipage}
        \begin{minipage}[t]{\linewidth}
            \includegraphics[width=\textwidth]{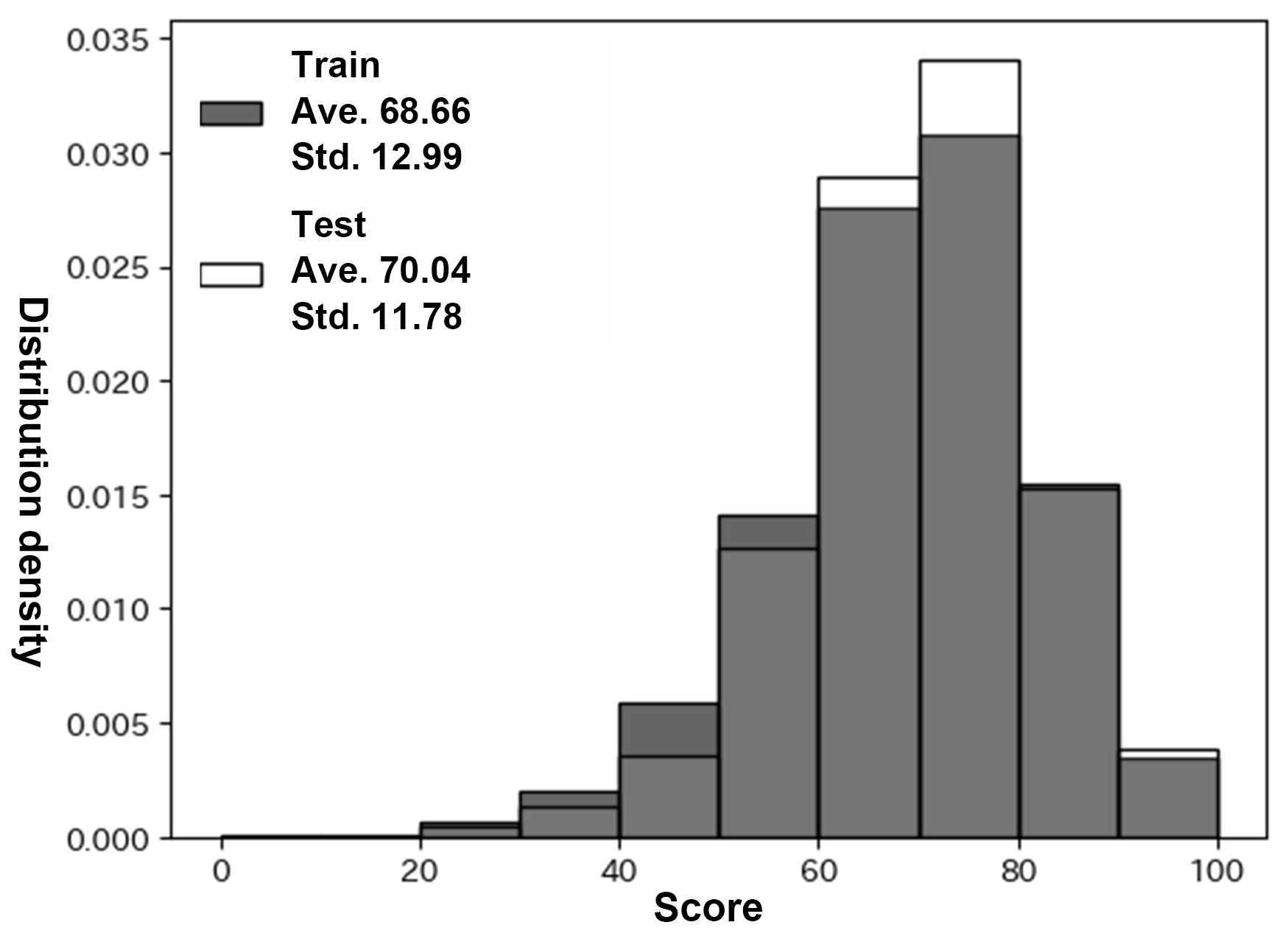}
            \caption{Label distribution of \taskpe.}
            \label{fig:label-distribution-pe}
        \end{minipage}
    \end{minipage}
\end{figure*}

\section{Additional Information in Experiment}\label{sec:exp-add}

\subsection{Model Details}

The versions and URLs of the models we used are listed in Table \ref{tab:model-names}. The detailed settings for the pretraining of fine-tuned models, such as the tokenizer and dataset, are provided in Table \ref{tab:model-detail}.

\subsection{Hyperparameters}
We used PyTorch~\cite{pytorch} and HuggingFace~\cite{huggingface} libraries for the model implementation.
We present the hyperparameters employed during the training of both fine-tuned encoder models and zero-/few-shot settings on LLMs are detailed in Table \ref{tab:hyperparam}.

\begin{table*}[t]
\centering
\resizebox{\textwidth}{!}{%
\begin{tabular}{@{}lccccc@{}}
\toprule
\multirow{2}{*}{\textbf{Evaluator}}
    & \textbf{\textsc{Ad Accept.}}
    & \textbf{\textsc{Ad Consist.}}
    & \textbf{\textsc{Ad Perform. Est.}}
    & \textbf{\taskaaa}
    & \textbf{\tasksim}
    \\
    & \footnotesize{Accuracy/F1-score}
    & \footnotesize{Accuracy/F1-score}
    & \footnotesize{Pearson/Spearman}
    & \footnotesize{F1-micro/-macro}
    & \footnotesize{Pearson/Spearman}
    \\
\midrule
\multicolumn{6}{c}{\textit{Fine-tuned Encoder Models}} \\[0.1cm]
Tohoku BERT$_{\rm BASE}$  & 0.685/0.691                   & 0.757/0.504          & 0.437/0.454                   & 0.753/0.629          & 0.769/0.803                   \\
Tohoku BERT$_{\rm LARGE}$ & 0.711/0.688                   & \textbf{0.767}/0.552 & \textbf{0.480}/\textbf{0.497} & 0.774/\textbf{0.694} & 0.773/0.807                   \\
Waseda BERT$_{\rm BASE}$  & 0.615/0.639                   & 0.725/0.388          & 0.444/0.454                   & 0.641/0.442          & 0.749/0.797                   \\
Waseda BERT$_{\rm LARGE}$ & 0.598/0.637                   & 0.755/0.474          & 0.445/0.457                   & 0.663/0.517          & 0.740/0.800                   \\
XLM-RoBERTa$_{\rm BASE}$  & 0.694/0.677                   & 0.743/0.465          & 0.425/0.439                   & 0.730/0.542          & 0.846/0.870                   \\
XLM-RoBERTa$_{\rm LARGE}$ & 0.705/0.690                   & 0.758/0.519          & 0.453/0.457                   & \textbf{0.778}/0.677 & \textbf{0.878}/\textbf{0.878} \\
\cmidrule{1-6}
\multicolumn{6}{c}{\textit{Zero-/Few-shot LLMs}} \\[0.1cm]
\calm                     & 0.520/0.115                   & 0.381/0.472          & 0.006/0.013                   & 0.154/0.042          & 0.036/0.036                   \\
\elyza                    & 0.352/0.520                   & 0.628/0.771          & 0.003/0.046                   & 0.196/0.044          & 0.015/-0.004                  \\
GPT-3.5                   & 0.369/0.489                   & 0.528/0.570          & -0.013/-0.022                 & 0.255/0.064          & 0.389/0.385                   \\
GPT-4                     & 0.325/0.433                   & 0.583/0.612          & 0.028/0.073                   & 0.417/0.113          & 0.776/0.811                   \\
\cmidrule{1-6}
\multicolumn{6}{c}{\textit{Fine-tuned LLMs}} \\[0.1cm]
\elyza      & 0.638/0.638 & 0.692/0.694          & 0.240/0.235                   & 0.379/0.280          & 0.684/0.740                  \\
\cmidrule{1-6}\morecmidrules\cmidrule{1-6}
Human                     & \textbf{0.732}/\textbf{0.790} & 0.703/\textbf{0.807} & ---                           & 0.564/0.538          & 0.699/0.765                   \\
\bottomrule
\end{tabular}%
}
\caption{Complete version of Table \ref{tab:eval-res} which showing the performance of PLMs and human evaluators on the test set. Bold indicates the best result across all methods.}
\label{tab:eval-res-full}
\end{table*}

\subsection{Prompts}
The template for the prompt in each task is illustrated in Figure \ref{fig:llm-prompt-template}.
We begin by outlining the purpose and overview of the task, followed by an explanation of the criteria and standards for evaluation. Next, we provide instructions on the response format, and in some cases, additional few-shot examples are included.
Tables \ref{tab:llm-prompt-1} and \ref{tab:llm-prompt-2} show the prompts actually used in the tasks.
Terms in curly brackets are variables to be filled in.
All variables, except \texttt{\{few\_shot\_examples\}}, correspond to the fields described in Section \ref{sec:task-design}.
Note that all prompts are translated from Japanese into English for presentation brevity.

\begin{table*}[t]
    \centering
    \resizebox{0.85\textwidth}{!}{%
        \begin{tabular}{l}
        \toprule
            \multicolumn{1}{c}{\taskaccept} \\[0.1cm]
Based on the following settings, determine whether the input sentence is acceptable as an advertisement expression.\\
\\
\# Criteria\\
- Output \textit{acceptable} if the sentence is coherent, fluent, and easy to read.\\
- Output \textit{unacceptable} if the sentence is unnatural due to unnecessary and excessive repetition, \\
inappropriate use of symbols, etc.\\
\\
\# Formats\\
- Answer with either \textit{acceptable} or \textit{unacceptable}.\\
- Do not output anything other than your answer.\\
\\
\texttt{\{few\_shot\_examples\}}\\
Input sentence: \texttt{\{ad\_text\}}\\
Output:\\
\cmidrule{1-1}
\multicolumn{1}{c}{\taskconsist} \\[0.1cm]
Based on the following settings, determine whether the given input sentence is consistent with the LP text.\\
\\
\# Criteria\\
- Output \textit{inconsistent} if the input sentence contains expressions not included in the LP text.\\
- Output \textit{inconsistent} if the input sentence outputs different numbers or names from those mentioned in the LP text.\\
\\
\# Formats\\
- Answer with either \textit{consistent} or \textit{inconsistent}.\\
- Do not output anything other than your answer.\\
\\
\texttt{\{few\_shot\_examples\}}\\
Input sentence: \texttt{\{ad\_text\}}\\
LP text: \texttt{\{lp\_text\}}\\
Output:\\
\cmidrule{1-1}
\multicolumn{1}{c}{\taskpe} \\[0.1cm]
Based on the following settings, estimate the performance of the given advertisement information.\\
\\
\# Criteria\\
- The good quality of the ad text (whether it is attractive, whether it is effective in the industry)\\
- The relevance of keywords and ad text\\
- The relevance of LP text and ad text\\
\\
\# Formats\\
- Do not output anything other than your answer.\\
- Answer with a number between 0 and 100.\\
\\
\texttt{\{few\_shot\_examples\}}\\
Industry: \texttt{\{industry\_type\}}\\
Keyword: \texttt{\{keyword\}}\\
Title: \texttt{\{title\_1\}} \texttt{\{title\_2\}} \texttt{\{title\_3\}}\\
Description: \texttt{\{description\_1\}} \texttt{\{description\_2\}}\\
Output:\\
\cmidrule{1-1}
\multicolumn{1}{c}{\tasksim} \\[0.1cm]
Based on the following settings, rate the advertising similarity of the given two input sentences.\\
\\
\# Criteria\\
- Judge based on the similarity in product category/product name and the similarity in appeal axis/persuasive \\
expressions.\\
- Closer to 5 if similar, closer to 1 if not similar.\\
\\
\# Formats\\
- Do not output anything other than your answer.\\
- Answer with a decimal number between 1 and 5.\\
\\
\texttt{\{few\_shot\_examples\}}\\
Input sentence 1: \texttt{\{ad\_text\_1\}}\\
Input sentence 2: \texttt{\{ad\_text\_2\}}\\
Output:\\
        \bottomrule
        \end{tabular}
    }
    \caption{Prompts used in \taskaccept, \taskconsist, \taskpe, and \tasksim~tasks.}
    \label{tab:llm-prompt-1}
\end{table*}

\ja{
\begin{table*}[t]
    \centering
    \resizebox{0.95\textwidth}{!}{%
        \begin{tabular}{l}
        \toprule
\multicolumn{1}{c}{\taskaaa} \\[0.1cm]
Based on the following settings, list all the aspects of advertising appeals included in the given input sentence.\\
Choose aspect labels from the following list. The list is in the format of "Label": "Description".\\
\\
\# List of aspects of advertising appeals\\
- \textit{Special deals}: Expressions emphasizing a sense of value such as price or discounts\\
- \textit{Discount price}: Expressions emphasizing specific prices or discounts\\
- \textit{Reward points}: Expressions emphasizing the rebate of points or money\\
- \textit{Free}: Expressions emphasizing that something is free\\
- \textit{Special gift}: Expressions emphasizing that perks are included\\
- \textit{Features}: Expressions emphasizing the content or features of the service\\
- \textit{Quality}: Expressions emphasizing high quality or a high-grade feel\\
- \textit{Problem solving}: Expressions emphasizing the solution to customers' problems\\
- \textit{Speed}: Expressions emphasizing the speed of delivery or procedures\\
- \textit{User-friendliness}: Expressions emphasizing the ease of use of the service\\
- \textit{Transportation}: Expressions emphasizing good accessibility\\
- \textit{Limited offers}: Expressions emphasizing some form of limitation\\
- \textit{Limited time}: Expressions emphasizing a limited period for service provision\\
- \textit{Limited target}: Expressions emphasizing that the service is provided to/for a limited target\\
- \textit{First-time limited}: Expressions emphasizing that the service is provided only for the first time\\
- \textit{Performance}: Expressions emphasizing the achievements of the service or company\\
- \textit{Largest/No.1}: Expressions emphasizing the scale or being No. 1 of the service or company\\
- \textit{Product lineup}: Expressions emphasizing the assortment or the number of stores\\
- \textit{Trend}: Expressions emphasizing that it is trending or in vogue\\
- \textit{Other}: Persuasive expressions suitable for advertising that do not fall into the above labels\\
- \textit{Story}: Expressions emphasizing the synopsis of the work\\
- \textit{No Match}: Label for when there is no persuasive expression\\
\\
\# Formats\\
- Output each aspects separated by ``|''\\
- Do not output anything other than your answer\\
- If no aspects apply, output ``\textit{No Match}''\\
\\
\texttt{\{few\_shot\_examples\}}\\
Input sentence: \texttt{\{ad\_text\}}\\
Output:\\
\bottomrule
        \end{tabular}
    }
    \caption{Prompt used in \taskaaa~task.}
    \label{tab:llm-prompt-2}
\end{table*}
}

\subsection{Evaluators} \label{sec:evaluators-add}
\paragraph{Fine-tuning Setting with Encoder Models.}
In the fine-tuning settings, models receive ad text and other values $\bm{x}=(x_i)^{|\bm{x}|}_{i=1}$ as input and predict labels $\bm{y}$, where $x_i$ represents the $i$-th token of the input sequence.
The hyperparameters used are shown in Appendix Table~\ref{tab:hyperparam}.
\taskaccept~and \taskconsist~involve predicting binary label $\bm{y} \in \{0, 1\}$ using an MLP from the encoded vector representation $h^{\mathtt{[CLS]}}$, where $\mathtt{[CLS]}$ is the special token for classification.
The inputs of the \taskaccept~task only use the ad text $\bm{x}^{\rm ad}$, while \taskconsist~uses $\bm{x}^{\rm ad} \oplus \bm{x}^{\rm LP}$ as input, where $\bm{x}^{\rm LP}$ and $\oplus$ represent the LP text and concatenation by a special token $\mathtt{[SEP]}$, respectively. \taskaaa~involves predicting all possible labels from a given ad text, $\bm{x}^{\rm ad}=(x^{\rm ad}_i)^{|\bm{x}^{\rm ad}|}_{i=1}$.
We adopted the doc-based architecture described in \citet{murakami2022aspect}, where the encoder model was used to obtain the vector representation $h^{\mathtt{[CLS]}}$.
This vector was then fed into an MLP layer, producing a label probability distribution $\bm{m} = \mathtt{Sigmoid}(\mathtt{MLP}(h^{\mathtt{[CLS]}})$, where $\bm{m}=(m_k)^K_{k=1}$ and $K$ is 21.
This number corresponds to the number of  \aaa~labels defined by \citet{murakami2022aspect}, as shown in Appendix Table \ref{tab:label-distribution}.
\tasksim~and \taskpe~are regression tasks that involve predicting a value of range $\bm{y} \in [1, 5]$ and $\bm{y} \in [0, 100]$, respectively.
Similar to the text classification task, we concatenated all inputs for each task with the special token $\mathtt{[SEP]}$, encoded the input into vector representation $h^{\mathtt{[CLS]}}$, and then fed the vector into the MLP layer to predict $\bm{y}$.

\paragraph{Zero-/Few-shot Setting with LLMs.}
Before conducting the experiment, we sampled 100 instances from the development set to adjust the parameters and prompts.
The prompts we used are shown in Figure \ref{fig:llm-prompt-template}.
Additionally, we determined the number of shots that performed best in the development set: 3-shot for \tasksim~and \taskpe~task, 2-shot for \taskaaa, and zero-shot for the others.
We calculated the final score by averaging five runs with different few-shot examples for a single instance.

\paragraph{Fine-tuning Setting with LLM.}
We fine-tuned ELYZA-7b, one of the open-sourced LLMs used in our study. We utilize QLoRA tuning, which applies the query and value of the attention module, training it for one epoch with the hyperparameters provided in Table \ref{tab:hyperparam}.

\subsection{Additional Results and Discussion on Fine-Tuning LLM Settings}

The results of this fine-tuning are detailed in Table \ref{tab:eval-res-full}. By fine-tuning ELYZA$_{\rm 7b}$, we observed performance improvements in most tasks as expected. Although it still falls short of fine-tuned encoder models, it outperformed GPT-4 in \taskaccept~and \taskconsist~tasks. While it did not match GPT-4's performance in \taskaaa~and \tasksim~tasks, it achieved performance levels close to human capabilities in \tasksim.

The extent of improvement varied across tasks. Notably, in \taskconsist, the accuracy increased by only 0.064, whereas in \tasksim, the Pearson and Spearman correlations improved dramatically to 0.659 and 0.736, respectively. The \taskaccept~and \taskaaa~tasks saw moderate improvements. This suggests that pre-training is particularly effective for tasks similar to \taskconsist, likely because the pre-training data, derived from web crawls, includes general knowledge applicable to \taskconsist. Conversely, the low scores in zero-/few-shot experiments for \tasksim~indicate that applying general knowledge from pre-training is challenging. However, fine-tuning appears to effectively map the pre-trained knowledge to the specific task, significantly enhancing performance. For \taskaccept~and \taskaaa~tasks, although additional training led to some improvement, the performance remained relatively low, suggesting that the provided training data was insufficient to fully grasp the knowledge required for these tasks.

\end{document}